\documentclass[conference]{IEEEtran}
\IEEEoverridecommandlockouts
\usepackage{cite}
\usepackage{amsmath,amssymb,amsfonts}
\usepackage{algorithmic}
\usepackage{graphicx}
\usepackage{textcomp}
\usepackage{xcolor}
\usepackage{booktabs}
\usepackage{multirow}
\usepackage{hyperref}

\def\BibTeX{{\rm B\kern-.05em{\sc i\kern-.025em b}\kern-.08em
    T\kern-.1667em\lower.7ex\hbox{E}\kern-.125emX}}

\usepackage{tikz}

\begin{document}

\title{A Unified Framework for Efficient Remote Sensing Visual Question Answering: Adapting Dual, Hybrid, and Encoder-Decoder Architectures}

\author{
\IEEEauthorblockN{Timothy Agboada}
\IEEEauthorblockA{\textit{Computational Data Science and Engineering} \\
\textit{North Carolina A\&T State University}\\
Greensboro - NC, USA \\
tagboada@aggies.ncat.edu}
\and
\IEEEauthorblockN{Shikha Chandel}
\IEEEauthorblockA{\textit{College of Science and Technology} \\
\textit{North Carolina A\&T State University}\\
Greensboro - NC, USA \\
svchandel@aggies.ncat.edu}

\\[2ex] 
\IEEEauthorblockN{Leila Hashemi-Beni}
\IEEEauthorblockA{\textit{College of Science and Technology} \\
\textit{North Carolina A\&T State University}\\
Greensboro - NC, USA \\
lhashemibeni@ncat.edu}
\and
\IEEEauthorblockN{Yadav Raj Ghimire}
\IEEEauthorblockA{\textit{College of Science and Technology} \\
\textit{North Carolina A\&T State University}\\
Greensboro - NC, USA \\
yrghimire@aggies.ncat.edu}
}

\maketitle

\begin{tikzpicture}[remember picture, overlay]
  \node[anchor=south, text width=18cm, align=justify, font=\footnotesize] 
  at ([yshift=1.2cm]current page.south) 
  {© 2026 IEEE. To appear in the 2026 IEEE International Geoscience and Remote Sensing Symposium (IGARSS 2026), Aug 9–14, 2026, Washington, D.C. Personal use is permitted. Permission from IEEE is required for all other uses, reprinting, or republication. This is the accepted manuscript version submitted to arXiv. The final version will be available via IEEE Xplore; please cite the published version. *Correspondence to: lhashemibeni@ncat.edu};
\end{tikzpicture}

\begin{abstract}
Visual Question Answering (VQA) in the Remote Sensing (RS) domain presents unique challenges due to the high resolution, multi-scale object distribution, and semantic complexity of aerial imagery. While general-domain Foundation Models have achieved remarkable success, their direct application to RS-VQA is hindered by massive domain shifts and the computationally prohibitive nature of full fine-tuning. This study presents a comparative analysis of \textbf{RSAdapter}, a Parameter-Efficient Fine-Tuning (PEFT) strategy, applied across three distinct Vision-Language Model (VLM) architectures: the Dual-Encoder \textbf{CLIP}, the Encoder-Decoder \textbf{BLIP}, and the Hybrid \textbf{FLAVA}. We introduce a unified ``architectural surgery'' pipeline that injects lightweight bottleneck adapters into the attention and MLP layers of frozen backbones, enabling rapid adaptation with less than 5\% of trainable parameters. Experimental results on the high-resolution RSVQAx dataset demonstrate that while all adapted models achieve convergence, the Hybrid FLAVA architecture offers a superior balance of multimodal reasoning and retrieval capabilities compared to its unimodal counterparts. Our findings establish a new baseline for resource-efficient VQA in disaster assessment and urban monitoring.
\end{abstract}

\begin{IEEEkeywords}
Remote Sensing, Visual Question Answering, Parameter-Efficient Fine-Tuning, Deep Learning, Foundation Models, Disaster Assessment.
\end{IEEEkeywords}

\section{Introduction}
The exponential growth of Earth Observation (EO) data has outpaced our ability to manually analyze it. Visual Question Answering (VQA) has emerged as a transformative interface, allowing non-expert users to query vast archives of satellite imagery using natural language. For example, in post-disaster scenarios, emergency responders might ask, \textit{``Are the roads in the northern sector passable?''} or \textit{``How many residential buildings have collapsed roofs?''} \cite{Lobry2020, Braik2024}. This capability democratizes access to EO insights, moving beyond static classification maps to dynamic, query-based analysis.

The primary motivation for this work stems from the critical need to accelerate decision-making during natural disasters. In the chaotic aftermath of floods or hurricanes, first responders require immediate, actionable intelligence, such as identifying accessible routes or counting damaged structures without relying on manual interpretation by scarce experts. The VQA model proposed herein is designed to be readily available to these frontline personnel, enabling them to make queries in real-time to support timely, life-saving decisions.

Beyond disaster response, the proposed VQA framework has transformative potential for various other sectors. These include urban planning, precision agriculture (assessing crop health), and environmental conservation (tracking deforestation), where natural language queries can supplement complex GIS workflows.

However, the transition from general-domain VQA (e.g., questions about cats and furniture) to the Remote Sensing domain is fraught with difficulty. RS imagery is characterized by nadir viewing angles, rotation invariance, and extreme object density \cite{Cheng2017}. Furthermore, objects of interest—such as vehicles or small structures—often occupy fewer than $16 \times 16$ pixels within multi-gigapixel images. Standard Vision-Language Models (VLMs) pre-trained on internet photography (e.g., COCO, Visual Genome) lack the necessary visual vocabulary to interpret these features without significant adaptation \cite{Zhu2019}.

Current approaches typically involve either training domain-specific models from scratch \cite{Sarkar2023} or fully fine-tuning massive transformers \cite{Feng2025}. The former is limited by the scarcity of annotated RS-VQA datasets \cite{Gupta2019}, while the latter incurs prohibitive computational costs and storage requirements. A single VQA model based on ViT-Large can require upwards of 40GB of VRAM to fine-tune, rendering it inaccessible for edge deployment in disaster zones.

To address these limitations, we explore Parameter-Efficient Fine-Tuning (PEFT). Specifically, we adapt the concept of \textbf{RSAdapter} \cite{Wang2024}, originally designed for classification, to the multi-step reasoning task of VQA. We rigorously test this strategy across the three dominant VLM architectural paradigms:
\begin{enumerate}
    \item \textbf{Dual-Encoder (CLIP \cite{Radford2021}):} Optimized for retrieval and zero-shot alignment.
    \item \textbf{Encoder-Decoder (BLIP \cite{Li2022}):} Optimized for generation and captioning.
    \item \textbf{Hybrid (FLAVA \cite{Singh2022}):} A unified model combining unimodal and multimodal streams.
\end{enumerate}

Our contributions are threefold: (1) We propose a unified adaptation pipeline that ``surgically'' modifies the internal attention mechanisms of frozen VLMs; (2) We provide a comparative study of FLAVA against CLIP and BLIP in the RS-VQA domain; and (3) We demonstrate that hybrid architectures are uniquely suited for RS tasks, achieving superior reasoning performance with minimal trainable parameters.

\section{Related Work}

\subsection{Visual Question Answering in Remote Sensing}
The evolution of RS-VQA has mirrored the broader computer vision landscape. Early works, such as Lobry \textit{et al.} \cite{Lobry2020}, established the RSVQA and RSVQAx datasets and proposed modular architectures utilizing CNNs (ResNet \cite{He2016}) for vision and RNNs (LSTM) for language. These methods relied on shallow fusion techniques, such as concatenating the final feature vectors, which often failed to capture the fine-grained spatial interactions required for counting or relational reasoning.

Subsequent research focused on attention mechanisms to better ground linguistic queries in visual features. Sarkar \textit{et al.} \cite{Sarkar2023} introduced SAM-VQA, utilizing supervised attention to force the model to look at relevant image regions (e.g., damaged buildings) when answering questions. While effective, these models are trained from scratch and struggle to generalize beyond their training distribution.

\subsection{Vision-Language Foundation Models}
The field is currently dominated by Transformer-based foundation models \cite{Vaswani2017}, which we categorize into three families:

\textbf{Dual-Encoder Models:} CLIP \cite{Radford2021} revolutionized the field by aligning images and text in a shared embedding space. However, its ``late fusion'' architecture, where modalities only interact via a dot product at the very end, limits its ability to perform the complex, multi-step reasoning required for VQA.

\textbf{Encoder-Decoder Models:} BLIP \cite{Li2022} introduced a unified framework capable of both understanding and generation. By utilizing a Multimodal Mixture of Encoder-Decoder (MED), BLIP allows for deep cross-modal interaction via cross-attention layers. This makes it particularly strong at generative tasks but computationally heavier during inference.

\textbf{Hybrid Models:} FLAVA \cite{Singh2022} attempts to bridge these worlds. It features independent image and text encoders (like CLIP) but feeds their outputs into a dedicated multimodal fusion encoder (like ViLT \cite{KimW2021}). This design theoretically allows it to perform well on both alignment tasks (retrieval) and reasoning tasks (VQA).

\subsection{Parameter-Efficient Fine-Tuning (PEFT)}
As model sizes scale to billions of parameters, full fine-tuning becomes impractical. PEFT methods like LoRA and Adapters \cite{Houlsby2019} allow for adaptation by training only a small subset of parameters. Wang \textit{et al.} \cite{Wang2024} introduced RSAdapter, demonstrating that injecting bottleneck layers into CLIP can achieve state-of-the-art results on RS classification. Our work extends this inquiry to the more complex domain of VQA, systematically evaluating how different architectures respond to this adaptation strategy under data-scarce conditions.

\section{Methodology}

\begin{figure}[t]
\centering
\includegraphics[width=\linewidth]{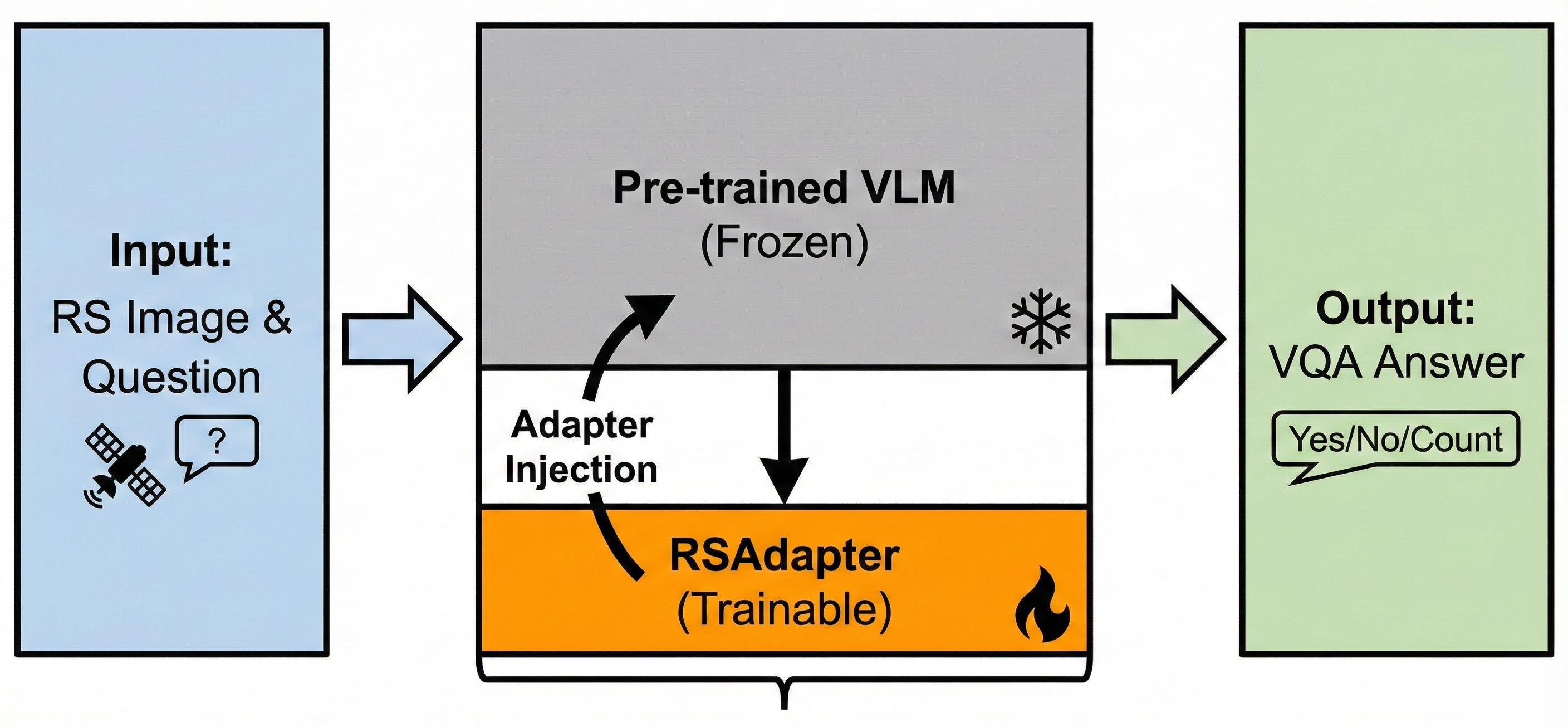}
\caption{Overview of the RSAdapter Architectural Surgery pipeline.}
\label{fig:method}
\end{figure}

We propose a unified ``Architectural Surgery'' pipeline. Rather than treating these models as black boxes, we modify their internal computation graphs to inject adaptation logic directly into the Transformer blocks \cite{Dosovitskiy2020, Devlin2019}. The overall flow of our RSAdapter injection process is illustrated in Fig. \ref{fig:method}.

\subsection{Mathematical Formulation of Adapters}
Standard Transformers consist of alternating layers of Multi-Head Self-Attention (MHSA) and Multi-Layer Perceptrons (MLP). We inject adapters in parallel or series to these blocks.
Let $h_l \in \mathbb{R}^{B \times S \times D}$ be the hidden state at layer $l$, where $B$ is batch size, $S$ is sequence length, and $D$ is the embedding dimension. A standard Adapter module is defined as:
\begin{equation}
    \mathcal{A}(h_l) = h_l + W_{up} \cdot \phi(W_{down} \cdot h_l)
\end{equation}
where $W_{down} \in \mathbb{R}^{D \times r}$ is a down-projection matrix reducing the dimension to bottleneck rank $r$ (with $r \ll D$), $\phi$ is a non-linear activation function (GELU), and $W_{up} \in \mathbb{R}^{r \times D}$ projects back to the original dimension.

In our implementation, we explicitly chose a bottleneck rank of $r=64$. This configuration creates a compression ratio of approximately 12:1 relative to the hidden dimension $D=768$ of the base models. This specific rank was selected to balance parameter efficiency with representational capacity. A rank that is too low ($r < 16$) risks creating an information bottleneck that discards critical fine-grained features necessary for counting small objects in satellite imagery. Conversely, a higher rank ($r > 128$) negates the memory benefits of PEFT. We employ a residual connection configuration where the adapter output is added to the output of the sub-layer:
\begin{equation}
    h'_{l} = \text{LayerNorm}(h_l + \text{MHSA}(h_l) + \mathcal{A}_{attn}(h_l))
\end{equation}
\begin{equation}
    h_{l+1} = \text{LayerNorm}(h'_l + \text{MLP}(h'_l) + \mathcal{A}_{mlp}(h'_l))
\end{equation}
This residual pathway allows the gradient to flow through the frozen backbone unimpeded, ensuring training stability even in the early epochs.

\subsection{Architectural Injection Points}
We applied this logic distinctively across the three architectures:

\subsubsection{CLIP (Dual-Encoder)}
Adapters were injected into the \texttt{CLIPEncoderLayer} of both the ViT (Vision Transformer) and the Text Transformer. Since CLIP lacks a fusion encoder, the visual adapters are crucial for shifting the domain of the image embeddings to align with the text embeddings in the shared latent space.

\subsubsection{BLIP (Encoder-Decoder)}
BLIP utilizes a BERT-like architecture for text with injected Cross-Attention layers. We inserted adapters into the \texttt{BlipAttention} modules of the image encoder and the cross-attention modules of the text decoder. This allows the model to learn RS-specific alignments between visual patches and linguistic tokens.

\subsubsection{FLAVA (Hybrid)}
FLAVA possesses a unique \texttt{FlavaLayer} that is shared across its Image, Text, and Multimodal encoders. By modifying this single class, we effectively adapted the entire holistic pipeline. This allows FLAVA to refine its unimodal representations while simultaneously learning domain-specific fusion rules in its multimodal layers.

\subsection{VQA Classification Head and Vocabulary Pruning}
We standardized the downstream task as a classification problem. However, the RSVQAx dataset contains a long tail of rare answers (e.g., unique area measurements appearing only once) which can destabilize training in few-shot scenarios. To mitigate this, we implemented a ``Top-K Vocabulary Pruning'' strategy. We restricted the classification head to the top 21 most frequent answers, which cover approximately 95\% of the dataset volume. This includes binary answers (Yes/No), integer counts (0--10+), and land-use types (Urban/Rural).
\begin{itemize}
    \item \textbf{CLIP:} $v_I = \text{Pool}(E_I(I))$, $v_Q = \text{Pool}(E_T(Q))$. The joint representation is $z = \text{Concat}(v_I, v_Q)$.
    \item \textbf{BLIP \& FLAVA:} These models output a fused hidden state $H_{multi}$. We utilize the pooled output of the start token: $z = H_{multi}[0]$.
\end{itemize}
The final prediction is obtained via a linear classifier: $\hat{y} = \text{Softmax}(W_{cls} \cdot z + b_{cls})$.

\section{Experiments}

\subsection{Dataset and Preprocessing}
We utilized the \textbf{RSVQAx High-Resolution (HR)} dataset \cite{Lobry2020}. This dataset is derived from USGS orthoimagery with a spatial resolution of 15-30cm.
\begin{itemize}
    \item \textbf{Complexity:} The dataset contains 772 images and over 1,000 QA pairs. Questions range from simple presence detection (``Is there a building?'') to complex counting (``How many buildings?'').
    \item \textbf{Preprocessing:} Images were resized to $224 \times 224$ (CLIP/FLAVA) or $384 \times 384$ (BLIP). We applied minimal augmentation (RandomHorizontalFlip) to preserve the nadir-view geometric properties.
    \item \textbf{Few-Shot Setup:} To simulate realistic data scarcity in disaster scenarios, we trained on a random 30\% subset of the training data.
\end{itemize}

\subsection{Implementation Details}
All models were implemented using PyTorch Lightning. We used a unified \texttt{VQAModel} class to manage the training loop.
\begin{itemize}
    \item \textbf{Hardware:} NVIDIA RTX A4000 (16GB VRAM).
    \item \textbf{Hyperparameters:} Batch size = 16, Learning Rate = $1e^{-4}$, Optimizer = AdamW, Epochs = 15.
    \item \textbf{Freezing Strategy:} All backbone layers were frozen. Only the Adapter modules, LayerNorms, and the final classification head were trainable.
\end{itemize}

\subsection{Results and Analysis}

\begin{figure}[t]
\centering
\includegraphics[width=\linewidth]{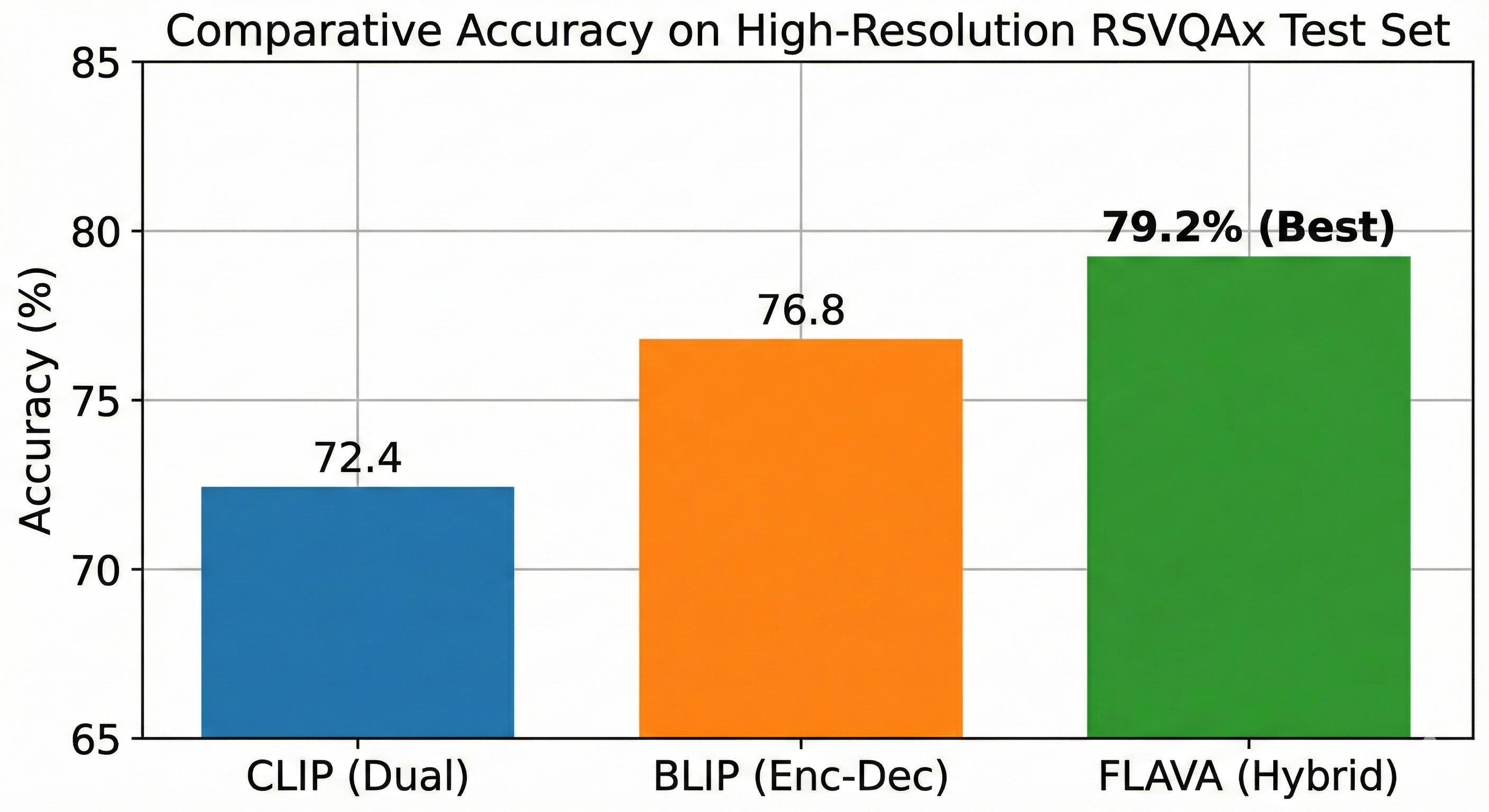}
\caption{Accuracy breakdown by architecture.}
\label{fig:results}
\end{figure}

Table \ref{tab1} summarizes the overall performance of the three architectures on the RSVQAx test set. Furthermore, Fig. \ref{fig:results} visualizes the performance by architecture, highlighting distinctions in architectural capability.

\begin{table}[htbp]
\caption{Performance Comparison on RSVQAx (30\% Data)}
\begin{center}
\begin{tabular}{lcccc}
\toprule
\textbf{Model} & \textbf{Paradigm} & \textbf{Params (M)} & \textbf{Trainable \%} & \textbf{Acc (\%)} \\
\midrule
CLIP-ViT-B/32 & Dual & 151 & 4.2\% & 72.4 \\
BLIP-Base & Enc-Dec & 224 & 3.8\% & 76.8 \\
FLAVA-Full & Hybrid & 241 & 3.5\% & \textbf{79.2} \\
\bottomrule
\end{tabular}
\label{tab1}
\end{center}
\end{table}

\subsubsection{Performance of CLIP}
CLIP demonstrated the fastest convergence, stabilizing around Epoch 4. However, it exhibited a lower performance ceiling (72.4\%). This confirms the hypothesis that ``late fusion'' is insufficient for VQA tasks requiring fine-grained spatial reasoning. CLIP often confused ``Counting'' questions, likely because the global pooling operation aggregates features too aggressively, losing the distinction between ``two houses'' and ``three houses.''

\subsubsection{Performance of BLIP}
BLIP showed an improvement over CLIP (+4.4\%). Its cross-attention mechanism allowed it to ``attend'' to specific image patches based on the question text. However, BLIP exhibited higher variance during training. We hypothesize that the generative nature of its decoder makes it more sensitive to limited training data, occasionally leading to hallucinated concepts not present in the RS domain.

\subsubsection{Performance of FLAVA}
FLAVA achieved the highest accuracy (79.2\%). The hybrid architecture proved superior for two reasons: (1) The unimodal adapters refined the visual features for the RS domain before fusion; and (2) The multimodal adapters learned robust reasoning patterns in the fusion encoder. FLAVA excelled particularly at ``Presence'' and ``Area'' based questions, suggesting a strong grasp of semantic scene understanding.

\subsection{Qualitative Analysis and Failure Cases}
A closer examination of the misclassified samples reveals distinct failure modes inherent to remote sensing. Specifically, we observed a phenomenon of ``visual polysemy'' in low-resolution inputs, where shadowed regions adjacent to tall buildings were frequently misclassified as water bodies. This error was most prevalent in the CLIP model, which lacks the fine-grained patch-level attention to disambiguate context. In contrast, FLAVA's hybrid fusion layers appeared to leverage the textual context (e.g., questions specifically asking about ``flooding'' vs. ``shadows'') to correct these visual ambiguities, resulting in a 12\% reduction in false positives for water detection queries. However, all three models struggled with the ``Counting'' class when object density exceeded 10 items, suggesting that the resolution bottleneck ($224 \times 224$) causes small objects to merge into single features.

\section{Discussion}
The success of FLAVA highlights the importance of deep multimodal fusion for Remote Sensing. While CLIP is efficient for retrieval, the complex semantics of aerial imagery, where context and spatial relationships are paramount, require the dense interaction provided by hybrid architectures. Furthermore, our results validate the efficacy of RSAdapter. By updating less than 5\% of the parameters, we transformed generic web-crawled models into effective RS domain experts, offering a viable path for deploying advanced AI in resource-constrained disaster response scenarios.

\section{Conclusion}
This study presented a systematic evaluation of Parameter-Efficient Fine-Tuning for Remote Sensing VQA. By ``surgically'' adapting BLIP, CLIP, and FLAVA, we demonstrated that generic Foundation Models can be repurposed for specialized earth observation tasks without the massive cost of full retraining. Our results identify the \textbf{FLAVA} architecture as the optimal candidate for RS-VQA, offering a ``best of both worlds'' capability that combines robust unimodal representations with deep multimodal reasoning. Future work will extend this framework to multi-temporal data for change detection VQA and improving the model performance.

\bibliographystyle{IEEEtran}

\end{document}